%% file: acl_latex.tex
\pdfoutput=1

\documentclass[11pt]{article}

\usepackage[final]{acl}

\usepackage{times}
\usepackage{latexsym}
\usepackage{amsmath}
\usepackage{amsthm}
\newtheorem{theorem}{Theorem}
\usepackage{amssymb}
\usepackage{dsfont}

\newenvironment{manualtheorem}[1]{
  \renewcommand{\thetheorem}{#1}
  \theorem 
}{
  \endtheorem 
}

\usepackage[T1]{fontenc}

\usepackage[utf8]{inputenc}

\usepackage{microtype}

\usepackage{inconsolata}

\usepackage{graphicx}
\usepackage{placeins}
\usepackage{algorithm}
\usepackage{algpseudocode}

\title{Quantifying Uncertainty in Natural Language Explanations of Large Language Models for Question Answering}

\author{Yangyi Li, Mengdi Huai \\
  Department of Computer Science,
  Iowa State University\\
  \texttt{\{liyangyi, mdhuai\}@iastate.edu} \\}

\begin{document}
\maketitle
\begin{abstract}
Large language models (LLMs) have shown strong capabilities, enabling concise, context-aware answers in question answering (QA) tasks. The lack of transparency in complex LLMs has inspired extensive research aimed at developing methods to explain large language behaviors. Among existing explanation methods, natural language explanations stand out due to their ability to explain LLMs in a self-explanatory manner and enable the understanding of model behaviors even when the models are closed-source. However, despite these promising advancements, there is no existing work studying how to provide valid uncertainty guarantees for these generated natural language explanations. Such uncertainty quantification is critical in understanding the confidence behind these explanations. Notably, generating valid uncertainty estimates for natural language explanations is particularly challenging due to the auto-regressive generation process of LLMs and the presence of noise in medical inquiries. To bridge this gap, in this work, we first propose a novel uncertainty estimation framework for these generated natural language explanations, which provides valid uncertainty guarantees in a post-hoc and model-agnostic manner. Additionally, we also design a novel robust uncertainty estimation method that maintains valid uncertainty guarantees even under noise. Extensive experiments on QA tasks demonstrate the desired performance of our methods.
\end{abstract}
\input{secs/1_introduction}
\input{secs/2_method}
\input{secs/3_experiment}
\input{secs/4_conclusion}
\input{secs/5_limitations}

\bibliography{anthology,custom}

\appendix

\input{secs/X_appendix}

\end{document}

%% file: secs/1_introduction.tex
\section{Introduction}
\label{sec:intro}

Large language models~(LLMs) such as GPT-4 have recently achieved impressive gains in natural‑language understanding and generation, demonstrating near‑human fluency across a wide range of tasks~\cite{achiam2023gpt}. When adapted for open‑domain question answering~(QA), these models exploit their vast parametric knowledge to deliver concise, context‑aware answers that surpass traditional retrieval pipelines in reasoning depth and coverage~\cite{kwiatkowski-etal-2019-natural,chen-etal-2025-benchmarking}. However, in LLM-based QA systems, the underlying LLMs are complex models where their inner working mechanisms are not yet fully understood. This lack of interpretability poses a significant barrier to their deployment in high-stakes decision-making applications, where inappropriate guidance can have severe consequences.

Given the importance of explaining LLMs' behaviors, many interpretation methods have been proposed \cite{zhu-etal-2024-explanation}. Among them, natural language explanations \cite{kumar-talukdar-2020-nile,camburu2018snli,wadhwa-etal-2024-learning} are appealing for their self-contained insights, even when the models are closed-source. However, there is no work studying rigorous uncertainties behind these explanations, making it difficult to understand the confidence level associated with them. Traditional uncertainty estimation methods (e.g., perturbation and Bayesian-based methods) \cite{tanneru2024quantifying,xiong2023can,liu-etal-2024-llms-learn-uncertainty} face significant limitations when applied to natural language explanations. Specifically, they either fail to provide valid uncertainty guarantees, or require the access of model logits and extensive model retraining.

In this work, we aim to provide rigorous, post‑hoc, and model‑agnostic uncertainty guarantees for natural‑language explanations in QA. Conformal prediction offers a distribution‑free framework with provably valid coverage \cite{shafer2008tutorial,su-etal-2024-api,10.1162/tacl_a_00715,qian2024towards,li2024data, lidder2025neuron, zhao2025membership,angelopoulos2024conformal}. However, traditional conformal prediction methods cannot be directly applied to natural language explanations. The reason is that in conformal prediction, the underlying models are typically trained in a supervised manner, where the uncertainty sets correspond to predefined class labels. In contrast, LLMs are trained in an auto-regressive fashion, generating text one token at a time, with each token conditioned on the previously generated tokens.

Additionally, real‑world QA queries often contain noise, such as ambiguous phrasing and typographical errors. Such noise can violate the underlying exchangeability assumption required by conformal prediction \cite{shafer2008tutorial}. Therefore, these generated uncertainties could become invalid, posing substantial challenges for generating reliable uncertainty estimates of natural language explanations. Although several robust conformal methods have been proposed \cite{yan2024provably,ghosh2023probabilistically,wang-etal-2024-conu,jeary2024verifiably}, they typically assume well-structured datasets and fail to account for discrete and token-level noise that is inherent in natural language explanations generated by LLMs. Such noise complicates valid uncertainty guarantees for natural language explanations in QA.

To address the above challenges, in this work, we propose \textbf{ULXMQA}, a novel \underline{\textbf{u}}ncertainty method for natural \underline{\textbf{l}}anguage e\underline{\textbf{x}}planations for \underline{\textbf{m}}edical \underline{\textbf{q}}uestion \underline{\textbf{a}}nswering, which can generate valid uncertainty guarantees in a post-hoc and model-agonistic way. Specifically, in our method, we first design prompts, which can assign each input token an importance score. Then, we design an uncertainty set construction function, which selects explanation tokens based on their assigned importance scores. For the constructed uncertainty sets, we provide theoretical guarantees by proving that the expected fraction of ground-truth tokens included in these uncertainty sets is theoretically guaranteed. Additionally, to address noisy data that may undermine the validity of the generated uncertainty sets, we also design a \underline{\textbf{r}}obust \underline{\textbf{u}}ncertain estimation method for these generated natural \underline{\textbf{l}}anguage e\underline{\textbf{x}}planations (\textbf{RULX}), which can provide robust valid uncertainty guarantees under discrete and token-level noise in questions. We further conduct extensive experiments to verify the desired performance of our proposed methods across different question answering tasks.

%% file: secs/2_method.tex
\section{Methodology}
\label{sec:method}
Here, we first introduce our valid uncertainty method for natural language explanations in LLM-based QA systems. Then, we present the proposed robust uncertainty method, designed to mitigate the effects of noise.

Without loss of generality, in this paper, we consider a vision-language model based QA system, which can output accurate answers to medical questions about the input medical image. Let \(\mathcal{Q}\) represent the question space and \(\mathcal{A}\) the answer space. We denote a language model as \(\mathcal{M}: \mathcal{Q} \rightarrow \mathcal{A}\), which takes a sequence of \(k\) question tokens \(Q = (q_1, q_2, \ldots, q_k)\), and produces a sequence of \(m\) answer tokens \(A = (a_1, a_2, \ldots, a_m)\). A designed prompt \(P\) augments the input and instructs $\mathcal{M}$ to emit a natural language explanation 
$E$, which we model as a subset of question tokens deemed essential for predicting $A$.

\textbf{Modeling uncertainty in natural language explanations.} Here, we propose a post‑hoc, model‑agnostic method that assigns provably valid uncertainty to natural language explanations $E$. Let \(\mathcal{D}^{\text{cal}} = \{(P_i, Q_i, E_i^*, A_i)\}_{i=1}^n\) denote the calibration data, and \(Q_{n+1} \) denote the test-time question. Here, $E_i^*$ denotes the ground‑truth explanation supplied by human annotators. Each example comes with a gold explanation sentence, and annotators mark the question tokens they judge essential in light of that sentence
\cite{aggarwal-etal-2021-explanations}. Specifically, we construct an uncertainty set of \(Q_{n+1} \) that provides theoretical guarantees on the inclusion ratio of ground-truth natural language explanations. The challenge in providing guarantees is that the language model \(\mathcal{M}\) generates text auto‑regressively. To address this, for each  question  $Q_i \in \{Q_{i}\}_{i=1}^{n+1}$, we propose a confidence-aware prompt \(P_{i}\), which enables \(\mathcal{M}\) to output an importance score $\mathcal{S}(P_i, q_{i,j};\mathcal{M})\in[0,1]$ for each word in $Q_i$ and simultaneously obtain the final answer. These scores reflect how essential each token is for predicting the final answer. Then, for \(Q_i\), we construct its uncertainty explanation set as follows 
\begin{align}
\label{eq:CPSet}
    & \mathcal{C}_\lambda(Q_i;\mathcal{M}) = \notag\\
    &\quad\{q_{i,j} \in Q_i: \mathcal{S}(P_i, q_{i,j};\mathcal{M})  \geq1-\lambda\},
\end{align}
where $\lambda$ is a parameter that increases the size of the prediction sets as its value grows. To obtain the importance score, we concatenate a prompt $P_i$ to the given
question $Q_i$ using the template:`` Read the question and
assign each
word an importance score...''. Crucially, our method preserves its coverage guarantee regardless of the quality of these scores. Such prompt-based explanations can effectively reveal the underlying reasoning process behind model predictions~\cite{parcalabescu-frank-2024-measuring,he2024harnessing,sudhi2024rag}. Notably, our method ensures valid coverage guarantees regardless of variations in the quality of these prompt-based results across different LLMs.

\begin{figure*}[!t]
    \centering
    \begin{minipage}[t]{0.48\textwidth}
        \centering        \includegraphics[width=\linewidth]{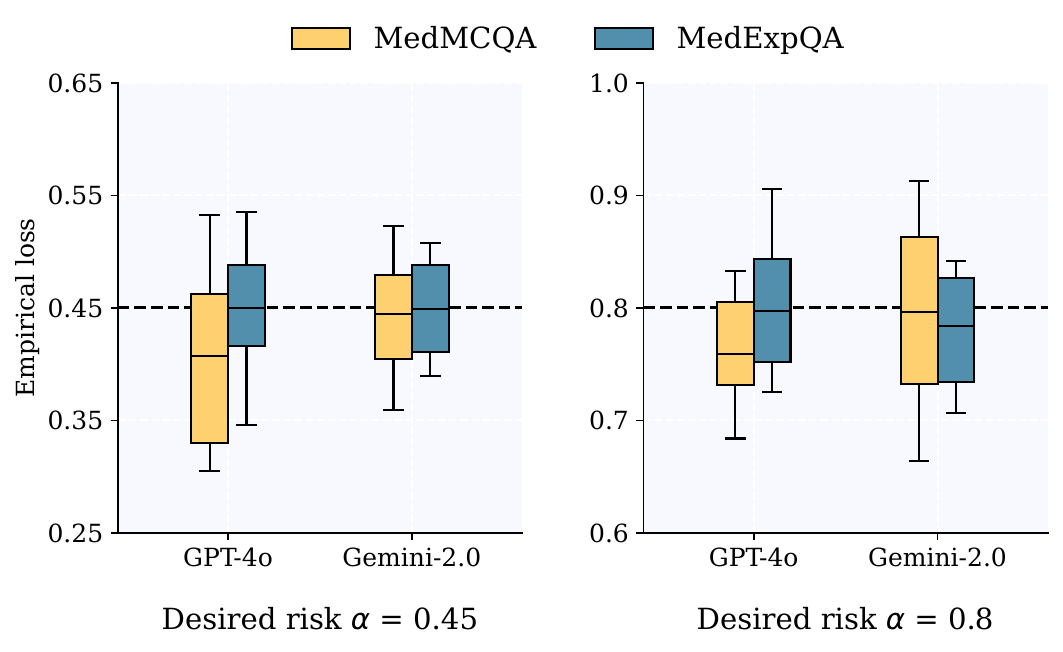}
        \caption{Validity of our ULXQA at desired risks.}
        \label{fig:validity_100data}
    \end{minipage}
    \hfill
    \begin{minipage}[t]{0.48\textwidth}
        \centering        \includegraphics[width=\linewidth]{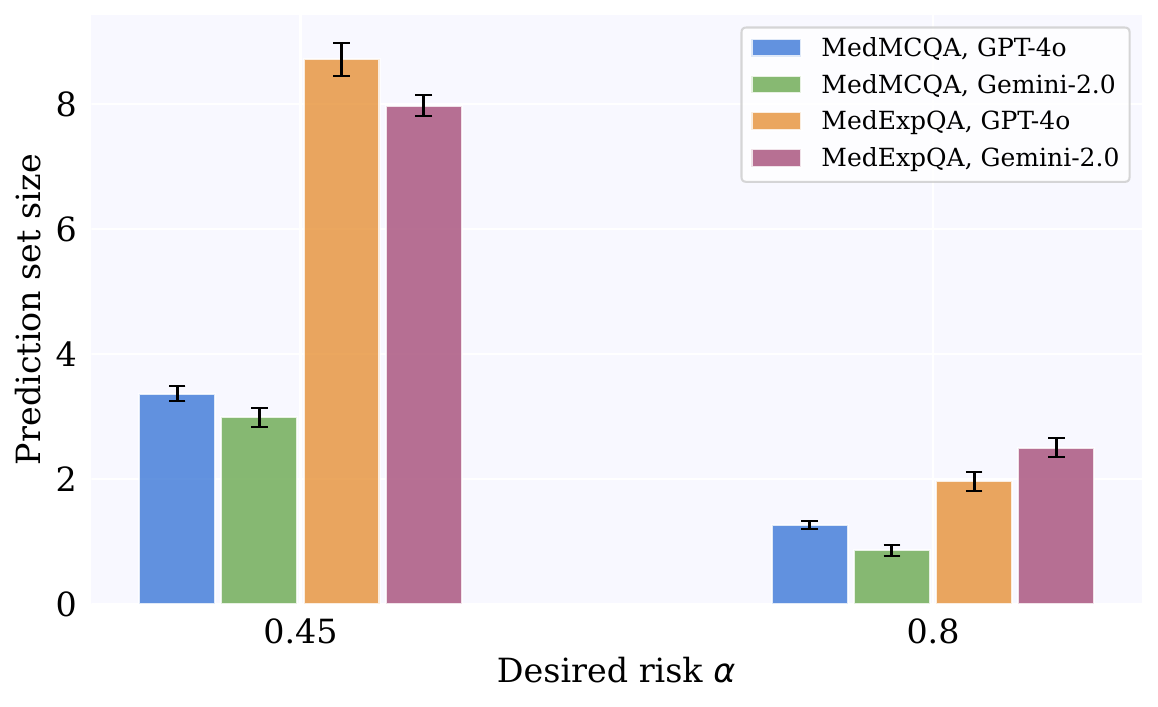}
        \caption{Efficiency of our ULXQA at desired risks.}
        \label{fig:efficiency}
    \end{minipage}
\vspace{-7pt}
\end{figure*}

To quantify the quality of the constructed  uncertainty set \(\mathcal{C}_\lambda(Q_i;\mathcal{M})\) of $Q_i$ from the calibration data $\mathcal{D}^{\text{cal}}$, we measure the proportion of ground-truth token explanations that appear in this set relative to the total ground-truth token explanations, and then define
\begin{align}
\label{eq:loss}
&\ell(\mathcal{C}_\lambda(Q_i;\mathcal{M}), E_i^{*},\lambda) \notag \\
&\quad =1-|E_i^{*}\cap\mathcal{C}_\lambda(Q_i;\mathcal{M})|/|E_i^{*}|.
\end{align} 
The above loss $\ell(\mathcal{C}_\lambda(Q_i;\mathcal{M}), E_i^{*},\lambda) $ decreases when the set $\mathcal{C}_\lambda(Q_i;\mathcal{M})$ includes a larger fraction of true tokens. Then, we calculate the average empirical loss at level $\lambda$ on the calibration set as $\widehat{R}_n(\lambda)=(\ell(\mathcal{C}_{\lambda}(Q_1;\mathcal{M}), E_1^{*},\lambda)+\ldots+\ell(\mathcal{C}_{\lambda}(Q_n;\mathcal{M}), E_n^{*},\lambda)) / n$. Given any desired risk level $\alpha \in (0,1)$, we set
\begin{align}
\label{eq:hat_lambda}
   \hat{\lambda}=\inf \{\lambda: \widehat{R}_n(\lambda) \leq \alpha-\frac{1-\alpha}{n}\}.
\end{align}
Since $\widehat{R}_n(\lambda)$ is monotone, we can efficiently search for $\hat{\lambda}$ using binary search to arbitrary precision 
and construct the uncertainty set \(\mathcal{C}_{\hat{\lambda}}(Q_{n+1};\mathcal{M})\) with uncertainty guarantees for the test-time question $Q_{n+1}$ based on Eq.~\eqref{eq:CPSet} and~\eqref{eq:hat_lambda}. The full algorithm is deferred to Algorithm 1 in the Appendix~\ref{sec:algo}.

\begin{theorem}
\label{thm:CRCThm} 
    Assume that the calibration set $\mathcal{D}^{\text{cal}}$ and the test data are exchangeable. For any desired $\alpha \in (0,1)$, let $\widehat{R}_n(\lambda)=(\ell(\mathcal{C}_{\lambda}(Q_1;\mathcal{M}), E_1^{*},\lambda)+\ldots+\ell(\mathcal{C}_{\lambda} \allowbreak(Q_n;\mathcal{M}), E_n^{*},\lambda)) / n$ and choose $\hat{\lambda}$ according to Eq.~\eqref{eq:hat_lambda}. Then, for the constructed uncertainty set \(\mathcal{C}_{\hat{\lambda}}(Q_{n+1};\mathcal{M})\), we have
    \begin{align}
    \label{eq:CRC_guarantee}
    &\mathbb{E}[\ell(\mathcal{C}_{\hat{\lambda}}(Q_{n+1};\mathcal{M}), E_{n+1}^{*},\hat{\lambda})] \leq \alpha.
    \end{align}
\end{theorem}
Theorem~\ref{thm:CRCThm} guarantees that, on average, the uncertainty set \(\mathcal{C}_{\hat{\lambda}}(Q_{n+1};\mathcal{M})\) contains at least a \(1-\alpha\) fraction of the true tokens, providing a valid coverage guarantee for the generated natural language explanations. Note that Theorem~\ref{thm:CRCThm} is stated under the exchangeability assumption, which is weaker than independence and identical distribution (i.i.d.). \emph{The proof of Theorem~\ref{thm:CRCThm} is provided in the Appendix~\ref{sec:thm}.}

\begin{figure*}[!t]
\centering
\begin{minipage}{0.48\textwidth}
\centering
\includegraphics[width=0.85\linewidth]{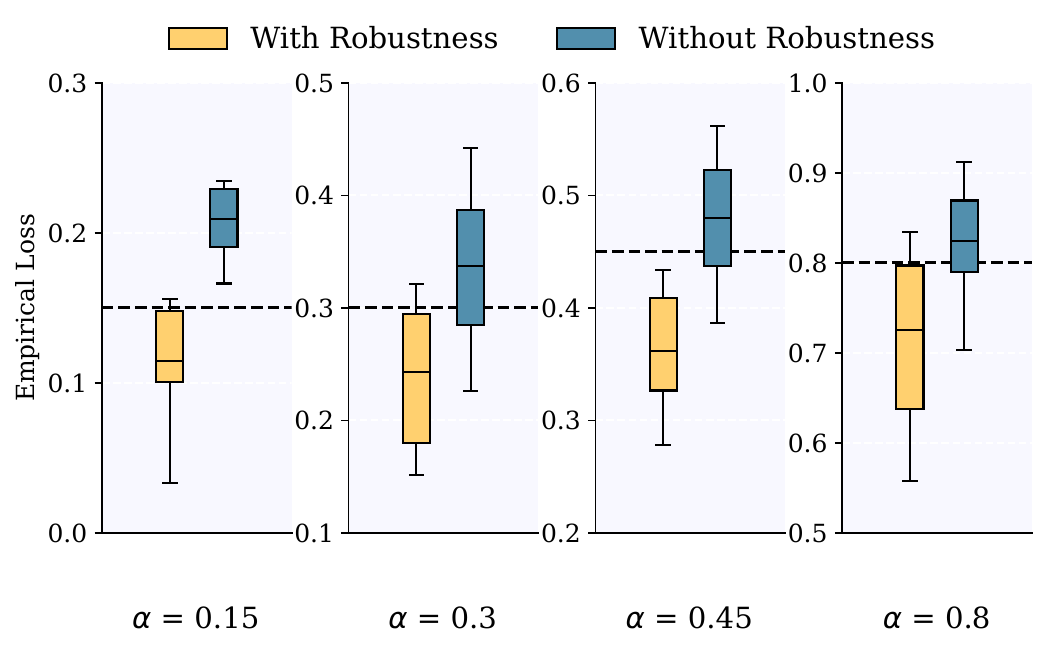}
\caption{Validity of our RULX on MedMCQA under noisy data.}
\label{fig:validity_100data_robust}
\end{minipage}
\hfill
\begin{minipage}{0.48\textwidth}
\centering
\includegraphics[width=0.85\linewidth]{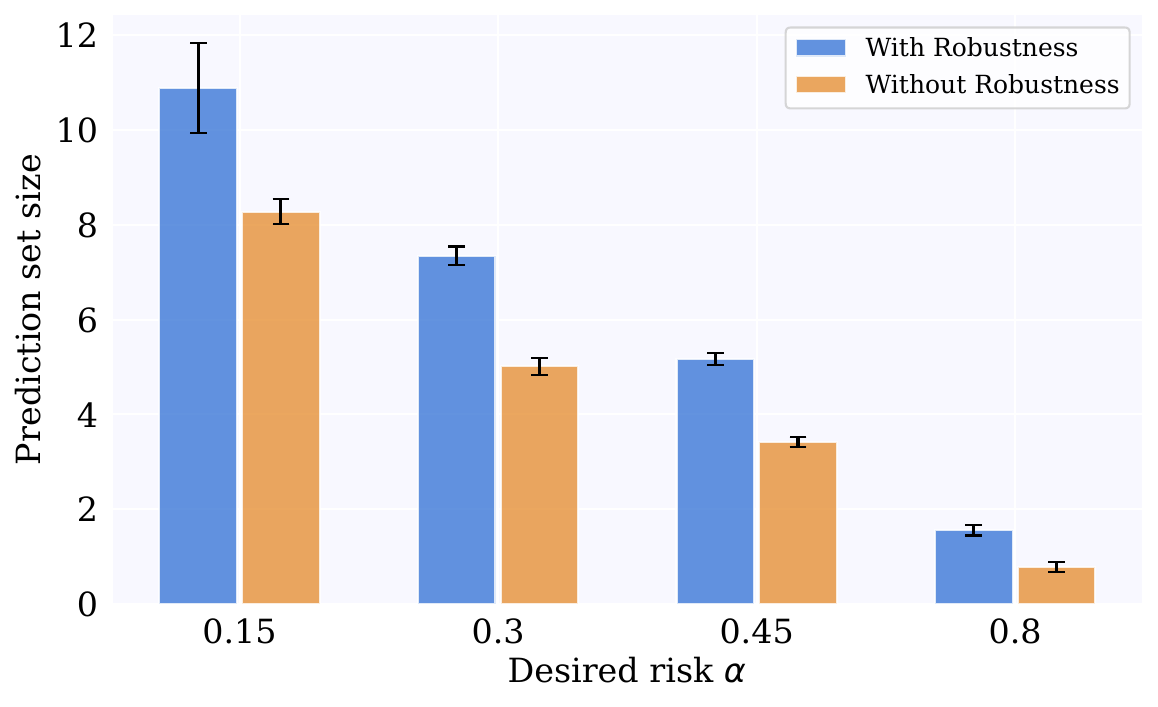}
\caption{Efficiency of our RULX on MedMCQA under noisy data.}
\label{fig:efficiency_robust}
\end{minipage}
\vspace{-10pt}
\end{figure*}

\textbf{Robust uncertainty guarantees under noisy data.} Note that the uncertainty guarantee of \(\mathcal{C}_{\hat{\lambda}}(Q_{n+1};\mathcal{M})\) relies on the exchangeability of the data,  which can be violated by noise such as ambiguous phrasing or typographical errors. Our goal is to provide robust uncertainty guarantees under noise. Let \(Q'_{n+1} \) represent the noisy test question derived from a clean version \( Q^*_{n+1} \). To model potential noise, we define \( \mathcal{B}_{Q^*_{n+1}} \) as the set of candidate noisy questions associated with \( Q^*_{n+1} \). As previously discussed, discrete and token-level noise inherent in natural language explanations poses significant challenges for robust uncertainties. To address this, for each word $q^*_{n+1,j} \in Q^*_{n+1}$, we define a synonym set $\mathcal{B}_{q^*_{n+1,j}}$, which contains the synonyms of $q^*_{n+1,j}$ (including $q^*_{n+1,j}$ itself).  Consequently, if noise affects at most $d \leq k$ words in $Q^*_{n+1}$, replacing them with elements from their respective synonym sets, the observed noisy question $Q'_{n+1} = \{q'_{n+1,1}, \cdots, q'_{n+1,k}\}$ emerges as follows 
\begin{align}
    \mathcal{B}_{Q^*_{n+1}} = &\{Q_{n+1}^{\prime}:\|Q_{n+1}^{\prime}-Q^*_{n+1}\|_0 \leq d, \\ 
    & \qquad q'_{n+1,j} \in \mathcal{B}_{q^*_{n+1,j}}, \forall j\}, \notag
\end{align}
where $\|Q_{n+1}^{\prime}-Q^*_{n+1}\|_0:= \sum_{j=1}^{k} \mathds{1} \{q'_{n+1,j} \neq q^*_{n+1,j}\}$. 
For the noisy test question $Q'_{n+1} \in \mathcal{B}_{Q^*_{n+1}}$, note that $\mathcal{B}_{Q'_{n+1}}$ also contains the clean test question $Q^*_{n+1}$ because noise affects at
most $d \leq k$ words in $Q^*_{n+1}$. To construct its robust uncertainty set, for each word $\tilde{q}_{n+1,j} \in \mathcal{B}_{q'_{n+1.j}}$  we compute
\begin{align}
\label{eq:robust_score}
&\mathcal{R}(P_{n+1}, \tilde{q}_{n+1,j}; \mathcal{B}_{Q'_{n+1}}, \mathcal{M})\\
&= 
\sup_{\substack{\tilde{Q}_{n+1} \in \mathcal{B}_{Q'_{n+1}}, \tilde{q}_{n+1,j} \in \tilde{Q}_{n+1}}} 
\mathcal{S}(P_{n+1}, \tilde{q}_{n+1,j}; \mathcal{M}), \notag
\end{align}
where the supremum is taken over all noisy questions $\tilde{Q}_{n+1}\in \mathcal{B}_{Q'_{n+1}}$ that still contain $\tilde{q}_{n+1,j}$. Intuitively, this ensures we capture the maximum importance score of $\tilde{q}_{n+1,j}$ across all relevant perturbations. With $\hat{\lambda}$ defined as in Eq.~\eqref{eq:hat_lambda}, we can construct the robust uncertainty set as follows
\begin{align}
&\quad \mathcal{C}^{\mathcal{R}}_{\hat{\lambda}}(Q'_{n+1};\mathcal{M}) = \{  \tilde{q}_{n+1,j} \in \mathcal{B}_{q'_{n+1}}:\\
&\mathcal{R}(P_{n+1}, \tilde{q}_{n+1,j}; \mathcal{B}_{Q'_{n+1}}, \mathcal{M}) \geq 1 - \hat{\lambda} \}. \notag
\end{align}
Here, robust uncertainty set \( \mathcal{C}^{\mathcal{R}}_{\hat{\lambda}}(Q'_{n+1};\mathcal{M}) \) includes all tokens whose maximum importance scores exceed the threshold $1-\hat{\lambda}$.

\begin{theorem}
\label{thm:CRC_robust_guarantee}
    Let $Q'_{n+1} \in \mathcal{B}_{Q^*_{n+1}}$ be a noisy test question near the clean test question $Q^*_{n+1}$ such that $\|Q'_{n+1} - Q^*_{n+1}\|_0 \leq d$. For the above constructed uncertainty set $\mathcal{C}^{\mathcal{R}}_{\hat{\lambda}}(Q'_{n+1};\mathcal{M})$, it satisfies
    \begin{align}    
    \label{eq:CRC_robust_guarantee}
    \mathbb{E}[\ell(\mathcal{C}^{\mathcal{R}}_{\hat{\lambda}}(Q'_{n+1};\mathcal{M}), E_{n+1}^{*},\hat{\lambda})] \leq \alpha.
    \end{align}
\end{theorem}
According to Theorem~\ref{thm:CRC_robust_guarantee}, for the noisy question \(Q'_{n+1}\), the expected proportion of true tokens included in the uncertainty set \(\mathcal{C}^{\mathcal{R}}_{\hat{\lambda}}(Q'_{n+1};\mathcal{M})\) is guaranteed to be at least $1-\alpha$. \emph{Due to the space limitations, the proof of Theorem~\ref{thm:CRC_robust_guarantee} and the full algorithm for RULX are provided in the Appendix~\ref{sec:thm}\&\ref{sec:algo}.}
Our framework could be generalized to the full conformal prediction setting~\cite{martinez2023approximating,chen2024modeling,blot2025automatically}, where machine unlearning techniques~\cite{zhao2023static,zhao2024rethinking,qian2023towards,qian2024exploring,qian2025towards,chen2025survey} could be explored to mitigate the associated high computational costs of retraining.

%% file: secs/3_experiment.tex
\section{Experiments}
\label{sec:Exp}
\subsection{Experimental Setup}
\textbf{Datasets and models.} We evaluate our approaches on two real‑world QA datasets: MedMCQA~\cite{pmlr-v174-pal22a}, a large‑scale dataset of 194k multiple‑choice questions, and MedExpQA~\cite{ALONSO2024102938}, a multilingual set of 622 clinical‑case questions. Note that both datasets include ground-truth explanations. Additionally, we adopt two advanced LLMs, GPT‑4o~\cite{gpt4o} and Gemini 2.0 Flash \cite{Google2024}, to ensure a thorough evaluation. Our code will be publicly released upon acceptance. 

\textbf{Implementation details.} For our experiments, we utilize LLMs with a temperature setting of 1. For each adopted QA dataset, we use its validation data as the test data, and partition its training data into 70\% fine-tuning data and 30\% calibration data. All experiments are run for 10 trials, and we report the averaged experimental results. 

\subsection{Experimental Results}
\textbf{Validity.} We evaluate ULXQA's validity on MedMCQA and MedExpQA across both LLMs, reporting empirical loss in Eq.~\eqref{eq:loss} at risk levels $\alpha = 0.45$ and $\alpha = 0.8$. Note that these LLMs are fine-tuned on the adopted QA datasets. In Fig.~\ref{fig:validity_100data}, the dashed line marks the desired risk, while the horizontal line in each box shows the average empirical loss. Our proposed ULXQA consistently maintains loss below or equal to $\alpha$, ensuring compliance with Eq.~\eqref{eq:CRC_guarantee}. For instance, with $\alpha = 0.45$, our proposed ULXQA achieves an empirical loss of $0.44$ on MedExpQA using Gemini 2.0 Flash. These results confirm that ULXQA can provide valid uncertainty guarantees on these generated explanations.

\textbf{Efficiency.} In Fig.~\ref{fig:efficiency}, we explore the efficiency of our proposed ULXQA across various LLMs. We report the average set size at the desired risk levels \(\alpha = 0.45\) and \(\alpha = 0.8\), ensuring consistency with the validity experiments. As depicted, our proposed ULXQA consistently provides explanation uncertainty sets with small sizes, so that it remains efficient when doctors use these explanation uncertainty sets to make decisions. For instance, on the adopted MedMCQA dataset, the average set size of our ULXQA using Gemini 2.0 Flash is approximately 2.99 at $\alpha=0.45$. It shows that, on average, fewer than three words can provide correct explanations for users with valid coverage guarantees. This indicates that ULXQA can efficiently output uncertainty quantification of natural language explanations for QA tasks. 

\begin{figure}[t]
\centering
\includegraphics[width=0.99\linewidth]{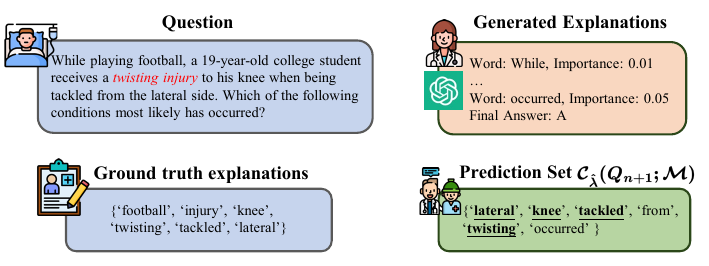}
\caption{Visualization results of our ULXQA on MedMCQA.}
\label{fig:visualization}
\vspace{-15pt}
\end{figure}

\textbf{Visualization.} We visualize uncertainty quantification for explanations on the MedMCQA dataset. In Fig.~\ref{fig:visualization}, the uncertainty set for a target image and question shows strong overlap with the ground truth, containing four correct words (`lateral', `knee', `tackled', and `twisting') in the prediction set. The prediction set size is six, which is close to the size of the ground truth. These results demonstrate our method’s ability to capture truly influential explanations with modest redundancy.

\textbf{Robust uncertainty under noisy data.} We evaluate RULX's validity and efficiency under varying risk levels using Gemini 2.0 Flash with noisy test data on MedMCQA. In Fig.~\ref{fig:validity_100data_robust}, the non-robust approach often exceeds desired risk levels, lacking formal guarantees, while RULX consistently stays within bounds, satisfying Eq.~\eqref{eq:CRC_robust_guarantee}. Fig.~\ref{fig:efficiency_robust} shows that although robust RULX slightly enlarges prediction sets to ensure validity, their sizes remain comparable to the non-robust method. Together, these figures confirm that RULX maintains valid uncertainty guarantees, while keeping prediction set sizes effectively comparable.

%% file: secs/4_conclusion.tex
\vspace{-2pt}
\section{Conclusion}
\vspace{-2pt}
\label{sec:Conclusion}
To the best of our knowledge, this work is the first to introduce a rigorous uncertainty estimation framework for natural language explanations in LLM-based QA systems. The post‑hoc, model‑agnostic method guarantees coverage by ensuring the expected fraction of non-ground-truth explanation tokens below a threshold. Building upon this, we further propose a robust extension that maintains reliable and valid uncertainty guarantees in the presence of noise. We also conduct extensive experiments across various QA tasks to comprehensively evaluate the effectiveness of our proposed methods.

\section*{Acknowledgments}
This work is supported in part by the US National Science Foundation under grants CNS-2350332 and IIS-2442750. Any opinions, findings, and conclusions or recommendations expressed in this material are those of the author(s) and do not necessarily reflect the views of the National Science Foundation.

%% file: secs/5_limitations.tex
\section*{Limitations}
\label{sec:limitations}
Our experiments show that ULXQA and its robust variant RULX achieve reliable coverage guarantees on challenging QA tasks, confirming the practical value of our framework. However, our current study has several limitations. First, the experimental results focus on the limited datasets, so additional experiments on other types of QA (e.g., legal or open‑domain) are needed to verify generality. Second, the present study considers only single‑modal textual inputs. An important next step is to extend ULXQA/RULX to multimodal settings, such as visual or audio question answering, and maintain uncertainty guarantees when multiple modalities are involved.

%% file: secs/X_appendix.tex
\section{Proofs of Theorems}
\label{sec:thm}
\begin{manualtheorem}{1}
        Assume that the calibration set $\mathcal{D}^{\text{cal}}$ and the test data are exchangeable. For any desired $\alpha \in (0,1)$, let $\widehat{R}_n(\lambda)=(\ell(\mathcal{C}_{\lambda}(Q_1;\mathcal{M}), E_1^{*},\lambda)+\ldots+\ell(\mathcal{C}_{\lambda} \allowbreak(Q_n;\mathcal{M}), E_n^{*},\lambda)) / n$ and choose $\hat{\lambda}$ according to Eq.~\eqref{eq:hat_lambda}. Then, for the constructed uncertainty set \(\mathcal{C}_{\hat{\lambda}}(Q_{n+1};\mathcal{M})\), we have
    \vspace{-5pt}
    \begin{align}
    \label{eq:CRC_guarantee_appendix}
    &\mathbb{E}[\ell(\mathcal{C}_{\hat{\lambda}}(Q_{n+1};\mathcal{M}), E_{n+1}^{*},\hat{\lambda})] \leq \alpha.
    \end{align}
\end{manualtheorem}
\begin{proof}
Consider a sequence of exchangeable random loss functions, \(\{\ell(\mathcal{C}_{\lambda}(Q_i;\mathcal{M}), E_i,\lambda)\}_{i=1}^{n+1}\), where \(\ell(\cdot,\cdot,\lambda)\) defined in Eq.~\eqref{eq:loss} is non-increasing in \(\lambda\), right-continuous, and satisfying \(\ell(\mathcal{C}_{\lambda_{\max}}(Q_1;\mathcal{M}),\cdot,\lambda_{\max})\leq \alpha\) when \(\lambda_{\max}=1\). We define
\begin{align}
    &\hat{R}_{n+1}(\lambda) = (\ell(\mathcal{C}_{\lambda}(Q_1;\mathcal{M}), E_1,\lambda)+\ldots \notag \\
    &\qquad +\ell(\mathcal{C}_{\lambda}(Q_{n+1};\mathcal{M}), \notag
    E_{n+1},\lambda))/(n+1),
\\
&\hat{\lambda}' = \inf\{ \lambda \in \Lambda : \hat{R}_{n+1}(\lambda) \leq \alpha \}.
\end{align}

Since $\inf_{\lambda} \ell =\ell(\mathcal{C}_{\lambda_{\max}}(Q_1;\mathcal{M}),\cdot,\lambda_{\max})\leq \alpha $, $\hat{\lambda}'$ is well-defined almost surely. Since $\ell(\mathcal{C}_{\lambda}(Q_{n+1};\mathcal{M}), E_{n+1},\lambda) \leq \sup_{\lambda} \ell = 1$, we get
\begin{align}
    &\hat{R}_{n+1}(\lambda) \notag \\
    &= \frac{n}{n+1} \hat{R}_n(\lambda) + \frac{\ell(\mathcal{C}_{\lambda}(Q_{n+1};\mathcal{M}), E_{n+1},\lambda)}{n+1} \notag \\
    &\leq \frac{n}{n+1} \hat{R}_n(\lambda) + \frac{1}{n+1}.
\end{align}

Thus, we can have
\begin{align}
    \frac{n}{n+1} \hat{R}_n(\lambda) + \frac{1}{n+1} \leq \alpha \Rightarrow \hat{R}_{n+1}(\lambda) \leq \alpha.\notag
\end{align}

This implies $\hat{\lambda}' \leq \hat{\lambda}$ when the LHS holds for some $\lambda \in [0,1]$. When the LHS is above $\alpha$ for all $\lambda \in [0,1]$, by definition, $\hat{\lambda} = \lambda_{\max} \geq \hat{\lambda}'$. Thus, $\hat{\lambda}' \leq \hat{\lambda}$ almost surely. Since $\ell(\cdot,\cdot,\lambda)$ is non-increasing in $\lambda$,
\begin{align}
\label{eq:expectless}
&\mathbb{E}[\ell(\mathcal{C}_{\hat{\lambda}}(Q_{n+1};\mathcal{M}), E_{n+1},\hat{\lambda})] \notag\\\leq &\mathbb{E}[\ell(\mathcal{C}_{\hat{\lambda}'}(Q_{n+1};\mathcal{M}), E_{n+1},\hat{\lambda}')]. 
\end{align}

Let $\mathcal{E}$ be the multiset of loss functions \(\{\ell(\mathcal{C}_{\lambda}(Q_i;\mathcal{M}), E_i,\lambda)\}_{i=1}^{n+1}\). Then $\hat{\lambda}'$ is a function of $\mathcal{E}$, or equivalently, $\hat{\lambda}'$ is a constant conditional on $\mathcal{E}$. Additionally, $\ell(\mathcal{C}_{\lambda}(Q_{n+1};\mathcal{M}), E_{n+1},\lambda) | \mathcal{E} \sim \text{Uniform}(\{\ell(\mathcal{C}_{\lambda}(Q_i;\mathcal{M}), E_i,\lambda)\}_{i=1}^{n+1})$ by exchangeability. These facts combined with the right-continuity of $L_i$ imply
\begin{align}
    &\mathbb{E} [ \ell(\mathcal{C}_{\hat{\lambda}'}(Q_{n+1};\mathcal{M}), E_{n+1},\hat{\lambda}') \mid\mathcal{E} ] \notag \\
&= \frac{1}{n+1} \sum_{i=1}^{n+1} L_i(\hat{\lambda}') \leq \alpha.
\end{align}

The proof is completed by the law of total expectation and Eq.~\eqref{eq:expectless}.
\end{proof}

\begin{algorithm}[t]
\caption{ULXQA}
\label{alg:ULXQA}
\begin{algorithmic}[1]
\Require A language model $\mathcal{M}$, importance score $\mathcal{S}(P_i, q_{i,j};\mathcal{M})$, calibration data $\mathcal{D}^{\text{cal}}$, test sample $Q_{n+1}$, Candidate threshold grid $\Lambda=\{\lambda_1<\lambda_2<\!\dots<\lambda_K\}$, desired risk level $\alpha \in (0,1)$
\Ensure Uncertainty set $\mathcal{C}_{\hat{\lambda}}(Q_{n+1}; \mathcal{M})$
\For{$i\gets1$ \textbf{to} $n$}
    \For{$k\gets1$ \textbf{to} $K$}
    \State Construct $\mathcal{C}_\lambda(Q_i;\mathcal{M}) =\{q_{i,j} \in Q_i: \mathcal{S}(P_i, q_{i,j};\mathcal{M})  \geq 1-\lambda\}$
    \State Compute $\ell(\mathcal{C}_\lambda(Q_i;\mathcal{M}), E_i^{*},\lambda_{k})$
    \EndFor
\EndFor
\State $low \gets 1,\; high \gets K$
\While{$low < high$}
    \State $mid  \gets  \bigl\lfloor\tfrac{low+high}{2}\bigr\rfloor$
    \State $\smash{\widehat{R}_n(\lambda_{mid}) \gets  \sum_{i=1}^{n} \frac{\ell(\mathcal{C}_\lambda(Q_i;\mathcal{M}), E_i^{*}, \lambda_{mid})}{n}
    }$    \If{$\widehat{R}_n(\lambda_{mid})\le \alpha$}
        \State $high \gets mid$  
    \Else
        \State $low \gets mid + 1$ 
    \EndIf
\EndWhile
\State $\hat{\lambda} \gets \lambda_{low}$
\State Construct the uncertainty set $\mathcal{C}_{\hat{\lambda}}(Q_{n+1};\mathcal{M}) = \{q_{n+1,j} \in Q_{n+1}: \mathcal{S}(P_{n+1}, q_{n+1,j};\mathcal{M})  \geq1-\hat{\lambda}\}$ that satisfies Eq.~\eqref{eq:CRC_guarantee}
\end{algorithmic}
\end{algorithm}

\begin{manualtheorem}{2}
Let $Q'_{n+1} \in \mathcal{B}_{Q^*_{n+1}}$ be a noisy test question near the clean test question $Q^*_{n+1}$ such that $\|Q'_{n+1} - Q^*_{n+1}\|_0 \leq d$. For the above constructed uncertainty set $\mathcal{C}^{\mathcal{R}}_{\hat{\lambda}}(Q'_{n+1};\mathcal{M})$, it satisfies
    \vspace{-10pt}
    \begin{align}    
    \label{eq:CRC_robust_guarantee_appendix}
    \mathbb{E}[\ell(\mathcal{C}^{\mathcal{R}}_{\hat{\lambda}}(Q'_{n+1};\mathcal{M}), E_{n+1}^{*},\hat{\lambda})] \leq \alpha.
    \end{align}
\end{manualtheorem}
\begin{proof}
According to the definition of robust score in Eq.~\eqref{eq:robust_score}, we have
\begin{align}
    &\mathcal{R}(P_{n+1}, q^*_{n+1,j};\mathcal{B}_{Q'_{n+1}},\mathcal{M}) \notag\\ \geq &\mathcal{S}(P_{n+1}, q^*_{n+1,j};\mathcal{M}), 
\end{align}
for any clean token $q^*_{n+1,j}$, which means
\begin{align}
    &q^*_{n+1,j} \in \mathcal{C}_{\hat{\lambda}}(Q^*_{n+1};\mathcal{M}) \notag\\
    \Rightarrow &q^*_{n+1,j} \in \mathcal{C}^{\mathcal{R}}_{\hat{\lambda}}(Q'_{n+1};\mathcal{M}). 
\end{align}
Consequently, we have 
\begin{align}    &\mathbb{E}[\ell(\mathcal{C}^{\mathcal{R}}_{\hat{\lambda}}(Q'_{n+1};\mathcal{M}), E_{n+1},\hat{\lambda})] \notag \\
\leq &\mathbb{E}[\ell(\mathcal{C}_{\hat{\lambda}}(Q^*_{n+1};\mathcal{M}), E_{n+1},\hat{\lambda})] \leq \alpha,
\end{align}
which directly implies the result stated in Eq.~\eqref{eq:CRC_robust_guarantee_appendix}.    
\end{proof}

\begin{algorithm}[!t]
\caption{RULX}
\label{alg:RULX}
\begin{algorithmic}[1]
\Require A language model $\mathcal{M}$, importance score $\mathcal{S}(P_i, q_{i,j};\mathcal{M})$, calibration data $\mathcal{D}^{\text{cal}}$, test sample $Q_{n+1}$, Candidate threshold grid $\Lambda=\{\lambda_1<\lambda_2<\!\dots<\lambda_K\}$, desired risk level $\alpha \in (0,1)$
\Ensure Uncertainty set $\mathcal{C}^{\mathcal{R}}_{\hat{\lambda}}(Q'_{n+1};\mathcal{M})$
\For{$i\gets1$ \textbf{to} $n$}
    \For{$k\gets1$ \textbf{to} $K$}
    \State Construct $\mathcal{C}_\lambda(Q_i;\mathcal{M}) =\{q_{i,j} \in Q_i: \mathcal{S}(P_i, q_{i,j};\mathcal{M})  \geq 1-\lambda\}$
    \State Compute $\ell(\mathcal{C}_\lambda(Q_i;\mathcal{M}), E_i^{*},\lambda_{k})$
    \EndFor
\EndFor
\State $low \gets 1,\; high \gets K$
\While{$low < high$}
    \State $mid  \gets  \bigl\lfloor\tfrac{low+high}{2}\bigr\rfloor$
    \State $\smash{\widehat{R}_n(\lambda_{mid}) \gets  \sum_{i=1}^{n} \frac{\ell(\mathcal{C}_\lambda(Q_i;\mathcal{M}), E_i^{*}, \lambda_{mid})}{n}
    }$
    \If{$\widehat{R}_n(\lambda_{mid})\le \alpha$}
        \State $high \gets mid$  
    \Else
        \State $low \gets mid + 1$ 
    \EndIf
\EndWhile
\State $\hat{\lambda} \gets \lambda_{low}$
\State Compute the robust importance score $\mathcal{R}(P_{n+1}, \tilde{q}_{n+1,j}; \mathcal{B}_{Q'_{n+1}}, \mathcal{M})$ based on Eq.~\eqref{eq:robust_score} 
\State Construct the robust uncertainty set $\mathcal{C}^{\mathcal{R}}_{\hat{\lambda}}(Q'_{n+1};\mathcal{M}) = \{  \tilde{q}_{n+1,j} \in \mathcal{B}_{q'_{n+1}}:\mathcal{R}(P_{n+1}, \tilde{q}_{n+1,j}; \mathcal{B}_{Q'_{n+1}}, \mathcal{M}) \geq 1 - \hat{\lambda} \}$ that satisifies Eq.~\eqref{eq:CRC_robust_guarantee}
\end{algorithmic}
\end{algorithm}

\section{Algorithms}
\label{sec:algo}
\subsection{Algorithm for ULXQA}
Algorithm~\ref{alg:ULXQA} describes how to construct the uncertainty set \(\mathcal{C}_{\hat{\lambda}}(Q_{n+1};\mathcal{M})\) with the valid uncertainty guarantees for the test-time question $Q_{n+1}$. Specifically, this set ensures that, on average, it contains at least a  \(1-\alpha\) fraction of the ground-truth explanation tokens. This yields a provably valid coverage guarantee for the natural language explanations generated by the model.

\subsection{Algorithm for RULX}
Algorithm~\ref{alg:RULX} outlines the construction of the uncertainty set \(\mathcal{C}_{\hat{\lambda}}(Q_{n+1};\mathcal{M})\). This method provides a robust coverage guarantee that the expected fraction of ground-truth explanation tokens included in the uncertainty set remains at least $1-\alpha$, even in the presence of input noise. As a result, it offers reliable uncertainty quantification for natural language explanations under possible perturbations.

\section{Datasets}
We adopt the following datasets:
\begin{itemize}
\item \textbf{MedMCQA}. Released under the MIT License for research use, this large‐scale benchmark provides expert‑verified multiple‑choice questions spanning cardiology, oncology, pediatrics, neurology, infectious diseases, and other medical specialties. The dataset is publicly available and contains no personally identifiable information. It is entirely in English, which facilitates evaluation of English‑language medical QA systems. Although questions center on clinical scenarios, demographic attributes of the underlying patient groups are not specified because such information is absent from the source data.

\item \textbf{MedExpQA}. Distributed under the CC BY‑NC 4.0 license for non‑commercial research, MedExpQA contains 622 clinician‑authored multiple‑choice questions, each paired with gold‑standard explanations written by medical professionals. The corpus covers four languages (English, Spanish, French, and Italian) and supports cross‑lingual assessment of answer correctness and explanation quality. It is publicly released and free of personally identifiable information. While items focus on clinical reasoning, demographic details of the represented populations are not included due to the nature of the data.
\end{itemize}

\section{AI Assistance in Writing}
During manuscript preparation, we employed an AI language assistant like GPT-o3 for copy‑editing tasks—namely, correcting grammar, smoothing phrasing, and improving overall readability. The model was not used to generate scientific ideas, analyses, or conclusions, and it played no role in shaping the study’s methodology or results. Its contribution was limited to language polishing, with all substantive content and final editorial decisions made exclusively by the human authors.